\pdfoutput=1

\documentclass[11pt]{article}

\usepackage{acl}

\usepackage{times}
\usepackage{latexsym}

\usepackage[T1]{fontenc}

\usepackage[utf8]{inputenc}

\usepackage{microtype}

\usepackage{float}
\usepackage{url}
\usepackage{bm}
\usepackage{amssymb}
\usepackage{dsfont}
\usepackage{multirow}
\usepackage{comment}
\usepackage{multicol}
\usepackage{multirow}
\usepackage{booktabs}
\usepackage{xspace}
\usepackage{cleveref}
\usepackage{tabularx}
\usepackage[linewidth=1pt]{mdframed}
\usepackage{graphicx}
\usepackage{subcaption}
\usepackage[graphicx]{realboxes}
\usepackage{algorithm}
\usepackage{algpseudocode}
\usepackage{xcolor}
\usepackage{soul}

\newcolumntype{P}[1]{>{\centering\arraybackslash}p{#1}}

\definecolor{c1}{HTML}{4e79a7}%
\definecolor{c2}{HTML}{f28e2b}%
\definecolor{c3}{HTML}{009E73}%
\definecolor{c4}{HTML}{56B4E9}%
\definecolor{c5}{HTML}{CC79A7}%
\definecolor{c6}{HTML}{E69F00}%
\definecolor{c7}{HTML}{844E4D}%
\definecolor{c8}{HTML}{2D512A}%

\definecolor{oorange}{HTML}{d95f02}
\definecolor{bblue}{HTML}{7570b3}
\definecolor{ggreen}{HTML}{1b9e77}
\definecolor{ppurple}{HTML}{e37fbb}
\definecolor{lgreen}{HTML}{9CD24A}
\definecolor{yyellow}{HTML}{FFD52D}
\definecolor{ggold}{HTML}{E1BC89}
\definecolor{ggray}{HTML}{AAAAAA}

\usepackage{listings}
\definecolor{dkgreen}{rgb}{0,0.6,0}
\definecolor{gray}{rgb}{0.5,0.5,0.5}
\definecolor{mauve}{rgb}{0.58,0,0.82}
\usepackage{tikz}

\title{Large Language Models are Inconsistent and Biased Evaluators}

\author{Rickard Stureborg$^{1,2}$ \hspace{4mm} 
        Dimitris Alikaniotis$^1$ \hspace{4mm} 
        Yoshi Suhara$^{3,}$\thanks{~~Work done while at Grammarly.} \\
  $^1$Grammarly~~~$^2$Duke University~~~$^3$NVIDIA\\
  \texttt{rickard.stureborg@duke.edu}\\
  \texttt{dimitrios.alikaniotis@grammarly.com}\\
  \texttt{ysuhara@nvidia.com}}

\begin{document}
\maketitle
\begin{abstract}
    The zero-shot capability of Large Language Models (LLMs) has enabled highly flexible, reference-free metrics for various tasks, making {\em LLM evaluators} common tools in NLP. However, the robustness of these LLM evaluators remains relatively understudied; existing work mainly pursued optimal performance in terms of correlating LLM scores with human expert scores. 
In this paper, we conduct a series of analyses using the SummEval dataset and confirm that LLMs are {\em biased} evaluators as they: (1) exhibit familiarity bias---a preference for text with lower perplexity, (2) show skewed and biased distributions of ratings, and (3) experience anchoring effects for multi-attribute judgments.
We also found that LLMs are {\em inconsistent} evaluators, showing low ``inter-sample'' agreement and sensitivity to prompt differences that are insignificant to human understanding of text quality.
Furthermore, we share recipes for configuring LLM evaluators to mitigate these limitations. Experimental results on the RoSE dataset demonstrate improvements over the state-of-the-art LLM evaluators.

\end{abstract}

\section{Introduction}

The advancement of NLP research has relied much on automatic evaluation to conduct quantitative analysis by comparing proposed and existing solutions for shared problems. The use cases for automatic evaluation are extensive, but most famously text generation tasks such as text summarization and machine translation, with classic evaluation metrics, including the family of ROUGE~\citep{lin-2004-rouge} and BLEU~\citep{papineni-etal-2002-bleu} scores, still widely in use today.

A core limitation of automatic evaluation is in developing new metrics and scaling them beyond limited benchmark datasets, primarily due to their common reliance on reference outputs. While there is a line of work in reference-free automatic evaluation metrics, it is known that it is less reliable than the current reference-based metrics~\citep{fabbri-etal-2021-summeval,deutsch-etal-2022-limitations}.
Large Language Models (LLMs) have proven useful in this domain due to their demonstrated high natural language understanding abilities and performance at adhering to instructions. 
Furthermore, with the powerful zero-shot capability, LLMs do not require reference texts and can generate scores directly from the system output. This has led to great interest in developing LLM-based automatic evaluation metrics~\citep{zheng2023judging,fu2023gptscore,lin2023llmeval,chiang-lee-2023-closer,chen2023exploring,wang2023chatgpt,liu2023geval,gao2023humanlike,shen2023large,luo2023chatgpt,chan2023chateval}; LLM evaluators (also known as LLM-as-a-judge) have become part of automatic evaluation for commonly used benchmarks for a variety of NLP tasks~\citep{li2024leveraging,huang2024empirical} including LLM benchmarks such as MT-Bench~\citep{zheng2023judging}. %

However, little is known about the robustness of these LLM evaluators. A few studies have looked deeper into this point~\citep{wang2023large,zheng2023judging,liu2023llms, li2024leveraging}; there is a need for further analysis into potential risks and failure points when using them, especially if used in sensitive applications.
Therefore, in this paper, we aim to study two important characteristics of the LLM evaluator, namely {\em bias} and {\em consistency}, in order to understand and share the limitations of LLM evaluators. 
To this end, we conduct extensive experiments using GPT-3.5 and GPT-4, which are commonly used as LLM evaluators, with various prompts and generation configurations on the summarization evaluation benchmarks SummEval and RoSE datasets.

In this paper, we quantitatively analyze biases in LLM evaluators, while linking the biased behaviors with those of humans.

First, we use the perplexity as a familiarity metric and analyze the relationship between the average perplexity and each rating returned by the LLM evaluator. We show that the average perplexity shows a descending trend as the score increases. The results support that LLM evaluators have {\em familiarity bias }~\cite{zajonc1968attitudinal}---LLM evaluators tend to develop a preference for texts simply because they are familiar with them.
Second, we explore scoring granularity and report that LLM evaluators exhibit score biases, including {\em round number bias}~\citet{thomas2009heuristics}, assigning some scores more frequently than others. 
Third, we report that LLM evaluators experience {\em anchoring effects}~\citep{tversky1974anchoring} when multiple labels are predicted in one output.

Then, we analyze the consistency of the LLM evaluator and  show that LLM evaluators significantly change their judgments for different samples, demonstrating significantly lower {\em inter-sample} agreement than human experts' inter-annotator agreement.
We also analyze LLM evaluators' inconsistent behaviors by changing the prompt configuration that should not affect the judgment.

Throughout analyzing these issues, we compiled findings into a set of recipes for LLM evaluators. We used the recipes to develop our new LLM evaluator and compared it with two existing LLM evaluators for text summarization. Experiment results on the RoSE dataset~\citep{liu-etal-2023-revisiting} show that our new LLM evaluator statistically significantly improves upon the state-of-the-art.

\section{Methodology}
\label{sec:methodology}
Analysis and results in this paper are the result of more than 560,000 generated outputs by LLMs.

\subsection{Datasets}
To investigate the performance of LLM-based evaluators, we test predictions on two main datasets.
We use SummEval~\citep{fabbri-etal-2021-summeval} as our development set, perform extensive analyses of LLM-based evaluators on this set, and then use RoSE~\citep{liu-etal-2023-revisiting} as an evaluation set for our case study comparing our system with the current SOTA LLM evaluator for summarization.

\subsubsection{SummEval}
Introduced by \citet{fabbri-etal-2021-summeval}, SummEval is a dataset of human annotated evaluations for automatically produced summaries for CNN/Daily Mail news articles.
The dataset annotates summaries on four dimensions: Coherence (collective quality of sentences in the summary), Consistency (factual alignment with the source), Fluency (quality of the individual sentences), and Relevance (well-selected content).
The dataset includes expert human judgments for 16 summaries produced by varying models on 100 articles over these four dimensions.

\subsubsection{RoSE}
RoSE \citep{liu-etal-2023-revisiting} is a benchmark of three datasets covering common summarization datasets: 
    CNN/Daily Mail News articles \citep{CNNDM}, 
    SAMSum dataset on chat dialogues \citep{SAMSum}, and 
    XSum containing extremely short abstractive summaries of text documents \citep{XSum}.
Annotations for RoSE are done to record recall of ``Atomic Content Units (ACU)'', which is a recall-like metric measuring how many of the atomic facts displayed within an article were captured by the summary.
We choose this benchmark due to its target labels very unlikely inclusion in any OpenAI model training given the time of its release,
the high quality labels they achieve through a novel method for multi-stage annotation,
and three domains to stress test our system on.

\subsubsection{Models}
We run our experiments in the analysis on a mix of GPT-3.5 (gpt-3.5-turbo-0301) and GPT-4 (gpt-4-0613). GPT-4 consistently outperforms GPT-3.5-Turbo. For the eventual test evaluation reported in \Cref{sec:case_study} on RoSE, we run previous work and our own approach using GPT-4-Turbo. Perplexity calculations are done using text-davinci-003 to match the LLM evaluator models as close as possible. We report our values against our own implementation of G-Eval to limit any potential differences in performance due to changes by OpenAI.

\subsubsection{Prompts}
Following \citet{stureborg-etal-2024-characterizing}, we use a slight variations on a prompt derived from \citet{GEval} to prompt LLMs for scores.
The full prompt we use is shown in \Cref{fig:system_prompt} and \Cref{fig:user_prompt}.
This prompt takes five input strings: \texttt{metric}, \texttt{metric\_definition}, \texttt{aspects}, \texttt{article}, and \texttt{summary}.
We replace \texttt{metric} with a name describing what dimension of analysis to focus on.
For SummEval, this is replaced with the string `Coherence' to investigate the first label, for example.
Further, \texttt{metric\_definition} is replaced with a written explanation of what the metric is meant to indicate, while \texttt{aspects} explains some broader considerations that are helpful in assessing the quality of a summary on this dimension.
Finally, \texttt{article}, and \texttt{summary} are replaced with the source document and summary for the models to make a prediction on.

\begin{figure}[t]
    \centering
    \begin{lstlisting}
You are the automatic summary evaluator of a writing editor:
- You consider an input document and a corresponding summary
- You evaluate the summary according to one important quality:
    1. {{metric}} (1-10) - {{metric_definition}}
- All ratings are between 1-10 where 1 is very poor and 10 is very good.
- Your evaluation should be critical and careful, and should closely match the ratings of experts. This evaluation is very important.
- Consider these aspects when evaluating:
    {{aspects}}

The user will give you both the article (document) and summary, and prompt you to provide an evaluation. Respond with your integer 1-10 score first, then a rationale.

Example:
    \end{lstlisting}
    \caption{\textbf{System text input for prompting chat-based LLMs to generate automatic evaluation scores in text summarization}.
    This prompting strategy is generalized to allow for use of evaluating any metric(s) of interest, whether multiple or just one.}
    \label{fig:system_prompt}
\end{figure}

\subsection{Evaluation Metrics}
The goal of automatic evaluation is to provide scores highly correlated with human judgments on the task at hand.
In our work, we primarily measure this through Kendall's $\tau$ correlation on scores produced for each label in SummEval (Coherence, Consistency, Fluency, Relevance), following the convention in other work on automatic evaluation of text summarization.

\section{Results and Analysis}
\label{sec:analysis}

In this section, we perform extensive analysis into the performance of LLM evaluators, we uncover several issues of bias and inconsistency with these systems, and propose potential solutions. 

\subsection{Familiarity Bias}

\begin{figure}[t]
   \centering
   \includegraphics[width=.48\textwidth]{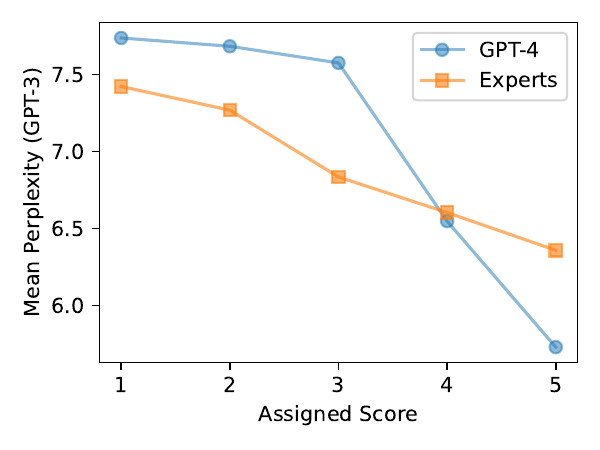}
   \caption{
       \textbf{Average perplexity for each rating by GPT-4 and Experts}.
       Summaries are grouped by evaluation scores (as assigned either by Experts or by GPT-4).
       GPT-4 exhibits a disproportionate bias toward low perplexity summaries compared to expert annotators, demonstrating a {\em familiarity bias}.
     }
    \label{fig:ppl_firstpage}
\end{figure}

We investigate the bias models have toward low perplexity examples.
Summaries are first grouped by evaluation scores (as assigned either by Experts or an LLM evaluator).
This group of summaries is held separate for each dimension of analysis in SummEval.
Perplexities are then computed with GPT-3 on the summary text, and a mean score is calculated for each group of summaries.
\Cref{fig:ppl_firstpage} shows that GPT-4 is disproportionately biased towards low perplexity summaries as
compared with expert annotators.
The mean perplexities of summaries assigned high scores (5s) are lower than that for expert raters, while mean perplexities of low assigned scores (1-3) are higher than expert raters.

Full results are reported in \Cref{tab:perplexity_analysis}. We would like to note that LLM evaluators are even biased by the source document, as LLM evaluators' ratings are still negatively correlated with the average perplexity of source documents, for which human experts' ratings show no correlation.
As system summaries in the SummEval are genearted by various summarization models and the perplexity of the summaries negatively correlates with the LLM evaluator's rating, we confirm that we can expand the notion of self-enhancement bias into familiarity bias.

\begin{table*}[t]
        \centering
        \scriptsize
        \begin{tabular}{r|rrrr|rrrr|rrrr|rrrr}
        \toprule
               & \multicolumn{8}{c|}{{\bf Avg. perplexity of summary}} & \multicolumn{8}{c}{{\bf Avg. perplexity of source document}} \\\midrule
               & \multicolumn{4}{c|}{GPT-4} & \multicolumn{4}{c|}{Human experts} & \multicolumn{4}{c|}{GPT-4} & \multicolumn{4}{c}{Human experts} \\
        Rating & Coh  &  Con &  Flu &  Rel &  Coh &  Con &  Flu & Rel  & Coh  &  Con &  Flu &  Rel &  Coh &  Con &  Flu & Rel \\\midrule
        1 &  --   & 7.05 &  --   & 8.42 & 7.03 & 7.47 & 7.66 & 7.53 &  --  & 7.76 &  --  & 8.51 & 7.21 & 7.66 & 7.68 & 7.94 \\
        2 & 8.15 & 7.61 & 7.45 & 7.53 & 6.80 & 7.42 & 7.71 & 7.14 & 8.32 & 7.86 & 7.77 & 7.79 & 7.67 & 7.61 & 8.01 & 7.53 \\
        3 & 7.60 & 7.46 & 7.92 & 7.33 & 6.58 & 7.07 & 6.96 & 6.73 & 8.09 & 7.69 & 8.48 & 7.90 & 7.73 & 7.52 & 7.55 & 7.67 \\
        4 & 6.44 & 6.83 & 6.48 & 6.44 & 6.37 & 6.67 & 6.96 & 6.42 & {7.72} & 8.00 & 7.74 & 7.75 & 7.81 & 7.29 & 7.99 & 7.81 \\
        5 & 5.34 & 6.06 & 6.01 & 5.51 & 6.36 & 6.43 & 6.39 & 6.26 & 6.51 & 7.44 & 7.06 & 6.84 & 7.63 & 7.75 & 7.69 & 7.58 \\
        \bottomrule
    \end{tabular}
    \caption{\textbf{Average Perplexity of Summary and Source documents for each rating by GPT-4/Human experts.}}    
    \label{tab:perplexity_analysis}
\end{table*}

\subsection{Scoring Granularity and Score Biases}
\label{sec:granularity}
A common scale for scoring is 1-5~\citep{nemoto2014likert}.
However, when producing scores for automatic evaluation, ties between candidate examples are often undesirable.
To reduce ties, we aim to increase scoring granularity: the distinct number of possible scores for candidate responses.
We explore the following methods for increasing granularity:
\begin{itemize}
  \setlength{\parskip}{0cm}
  \setlength{\itemsep}{0cm}
    \item 1-5 star: Resulting prediction when a model is instructed to provide an integer rating between 1-5 (inclusive).
    \item 1-5 + word modifier: Model is instructed to provide an integer rating between 1-5 along with a single word modifier indicating if it is `strong' or `weak'. 
    For example, a summary may be rated as a ``3'', ``weak 5'', or ``strong 4''.
    To map these ratings to a numerical value, we convert the `strong' modifier to add $0.33$ to the base rating, and `weak' subtracts 0.33 (similar to grading scales).
    \item 1-5 + float modifier: this score is directly predicting the resulting numerical value from the word modifier. 
    We instruct the model to predict values on a GPA scale (1.0, 1.33, 1.67, 2...).
    \item 1-10 score: instruct model to provide integer ratings between 1-10.
    \item 1-100 score: instruct model to provide integer ratings between 1-100.
\end{itemize}
For each of these cases, we also consider methods of taking a sample average.
In this approach, we produce N model responses 
    \footnote{For all experiments in this work, we set N=10 to balance reducing variance with avoiding prohibitive cost increases} 
    and average the resulting scores to provide a final float value with a greater granularity without changing the prompt.
This approach is similar to the approach outlined in G-Eval \citep{GEval}, where each potential score is multiplied by its token probability to get an expected value score.
Since OpenAI does not allow access to log probabilities of their top-end models we instead sample several times at a temperature of 1.0, which approximates the expected value score and maintains an increased granularity.

\begin{figure}[th]
   \centering
   \includegraphics[width=.48\textwidth]{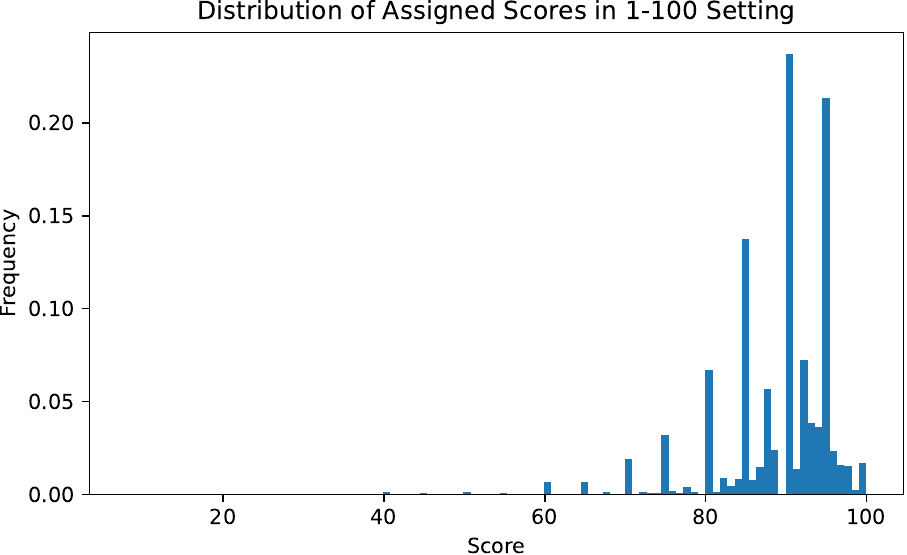}
   \caption{
       \textbf{Frequencies of each possible score as found in 64,000 predictions using the 1-100 scale}.
       Models sparsely predict scores within the range.
       Frequencies of some scores, such as 90 and 95, are far higher than `odd' scores such as 92 or 19, and much of the range is almost entirely ignored (1-60).
       Interestingly, 1-60 is a range often largely ignored in academic grading scales.
       This indicates an issue within instruction-following specific to automatic evaluation.
    }
    \label{fig:distr_100}
\end{figure}

\Cref{fig:distr_100} shows the distribution of scores produced when instructing GPT-3.5-Turbo and GPT-4 to rate summaries on a 1-100 scale.
The scores in this distribution are not respected as intended, and the model assigns outsized probabilities to certain scores such as 90 and 95.
This reaffirms results by \citet{zheng2023large} which found that multiple-choice selections by LLMs suffered from similar token biases, deteriorating performance.
The full range is also not utilized, with predicted scores largely occurring between 70 and 100.

\Cref{fig:distr_100} also shows that the score distribution has several peaks for round numbers such as 60, 70, 80, 90 (Similarly for 75, 85, and 95), indicating that LLM evaluators also have {\em round number bias} like human.\footnote{It is indeed an interesting open question that how LLMs inherit round number bias from human through text written by human. Existing work reported round number bias for human judgments~\citep{coupland2011frequent,honda2022round}.}

 \begin{table}[h]
        \centering
        \small
        \begin{tabular}{llrrrrr}
        \toprule
        \textbf{Method} & G & Coh & Con & Flu & Rel & \textbf{Avg} \\
        \midrule
        1-5 star         & 5   & .332 & .362 & .325 & .337 & .339 \\
        1-5 avg   & 41  & .422 & .370 & .356 & .439 & .397 \\
        5 +word mod.   & 13  & .361 & .408 & .345 & .363 & .369 \\
        5 +word (avg)  & 121 & .394 & .364 & .316 & .419 & .373 \\
        5 +float mod.  & 13  & .425 & \textbf{.453} & \textbf{.380} & .395 & .413 \\
        5 +float (avg) & 121 & .416 & .378 & .334 & .438 & .392 \\
        1-10 score       & 10  & .450 & .433 & .366 & \textbf{.462} & \textbf{.428} \\
        1-10 avg  & 91  & .424 & .366 & .332 & .435 & .389 \\
        1-100 score      & 100 & \textbf{.463} & .423 & .308 & .339 & .383 \\
        1-100 avg & 991 & .406 & .351 & .343 & .414 & .379 \\
        \bottomrule
    \end{tabular}
    \caption{\textbf{Correlation with human judgement for GPT-4 by method for increased granularity}.
    ``G'' is the effective granularity (number of unique scores) possible within the given scale.
    Methods denoted ``avg'' are a 10-sample average run with temperature 1.0, while all other methods benefited from reducing temperature to 0. 
    It seems that increasing granularity generally helps low-granularity methods, while high-granularity methods are harmed by increasing granularity.
    This may be due to the increase in temperature setting.
    Our results indicate that there may be diminishing returns of increasing scoring granularity.
    }    
    \label{tab:granularity_performances}
\end{table}

To verify which rating scales produce higher quality responses by LLM evaluator frameworks, we run a comparative analysis of the cases mentioned in \S\ref{sec:granularity}.
\Cref{tab:granularity_performances} shows performance of GPT-4 based evaluators on SummEval under the mentioned rating scales.
The performance of 1-10 score performs best on average, with an average score of 0.428 Kendall's $\tau$ across the labels in SummEval.
This method also performs the best on relevance, at 0.462 Kendall's $\tau$, while 1-100 scoring performs better on Coherence and the float modification method performs best on both Consistency and Fluency\footnote{It is unclear why the results should differ across each dimension, indicating another potential issue with LLM evaluation: hyper-parameters may not be stable across different labels.}.
Ultimately, increasing scoring granularity is shown to improve performance in our experiments, which should be carefully conducted for the risk of score bias and round number bias.

\subsection{Anchoring Effect in Multiple Judgments}
During evaluation of text, it is often helpful to describe several attributes regarding the text at the same time.
For some tasks (such as hierarchical classification~\citep{zhu2024hierarchical} or N-ary relation extraction~\citep{cheung2023polyie}), the large set of target labels and long required contexts make separating annotation into independent generations infeasible; it is cheaper to predict all labels within the same output~\citep{gao2023humanlike}.
We explore whether doing so is beneficial for the performance of the model, since it could be argued that this is similar to a multi-task setting where scores of one feature may help determine the correlation of others.
However, conditioning on previously generated scores may bias generation on previous predictions in the context, thereby worsening performance.

We prompt GPT-4 to produce scores for Coherence, Consistency, Fluency and Relevance in a single generation (in that order).
We then look at the distributions of, for example, Consistency given each predicted score on Coherence.
Formally, we are interested in using our predictions to estimate the conditional probability:\\
$P(\texttt{Consistency} = X \hspace{0.2em}| \hspace{0.2em}\texttt{Coherence} = Y)$

\begin{figure}[h]
   \centering
   \includegraphics[width=.48\textwidth]{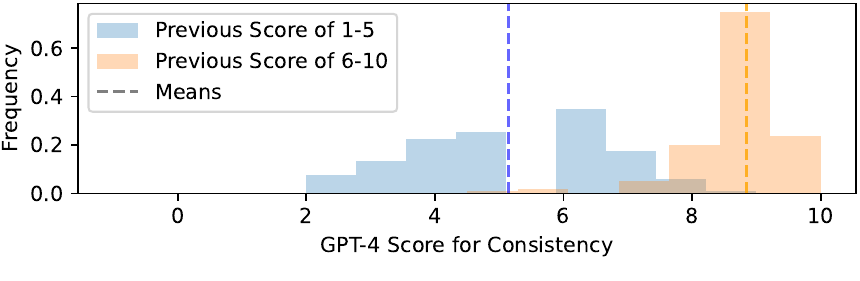}
   \includegraphics[width=.48\textwidth]{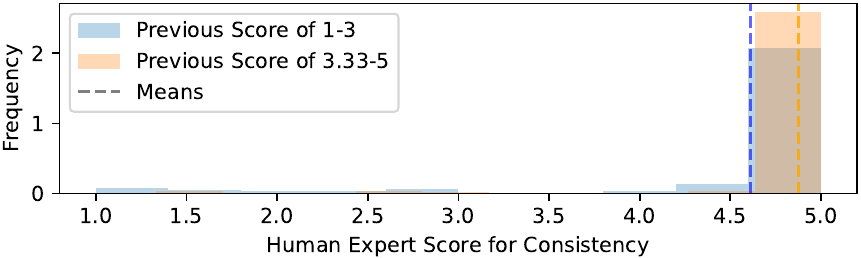}
   \caption{
       \textbf{(Top) Score distribution for \textit{consistency}, conditioned on the previously assigned score for \textit{coherence} when predicting both within the same context.
       (Bottom) Human-determined scores for \textit{consistency} conditioned on what range the score fell into for \textit{coherence}.}\footnotemark
       Human scores are correlated by  Pearson's $r = 0.315$, while GPT-4 scores are correlated by $r = 0.979$.
       The above figures clearly show how previous scores bias the distribution of future scores in the generation.
       While such biasing is natural (and in part valid), the effect here is so large it harms performance.
    }
    \label{fig:prob_of_con_given_coh}
\end{figure}

We then plot the frequency of evaluated scores when the previous score was above or below 5 out of 10.
\Cref{fig:prob_of_con_given_coh} shows one such plot, and the remainder of pairings are shown in \Cref{apx:prob_x_given_y}.
We find that there is a disproportionate biasing effect from the model, where the mean score assigned to samples with a previous assigned score above 5 is substantially greater than the mean score assigned to samples with previous scores of 5, while these scores should not be so strongly correlated.
In other words, LLM evaluators tend to overrely on this adjustment of its priors---experiencing an {\em anchoring effect}.
This is unsurprising due to LLM's auto-regressive generation, but points out the need to correct for such biases if utilizing multi-attribute predictions.

\footnotetext{Yet again we note the LLM evaluator does not make use of the full range of the scores, with no predictions of 5/10 for consistency in this experiment}

As seen in \Cref{fig:prob_of_con_given_coh} , one source of poor performance for GPT-4 is that humans mostly rate summaries as highly consistent (4-5) while GPT-4 questions consistency very often, assigning relatively low scores.

We run another experiment where we again generate all four scores on SummEval within one output, but change the relative order of the \textit{Coherence} attribute as compared to the other three attributes.

As in \Cref{tab:nth_element_performance}, we find labels predicted later in the LLM generation experience a degradation in correlation against expert annotators ($\tau$). The results indicate that the judgment for the target attribute (i.e., Coherence) was influenced by the previous judgments for the other attributes and LLM evaluators can experience {\em anchoring effects} when multiple attributes are judged in the same prompt.

\begin{table}[h]
        \centering
        \small
        \begin{tabular}{lrrrr}
        \toprule
        \textbf{N} & 1 & 2 & 3 & 4 \\
        \midrule
        $\tau$     & \textbf{0.400} & 0.391 & 0.359 & 0.368\\
        \bottomrule
    \end{tabular}
    \caption{\textbf{Performance of GPT-3.5-Turbo on \textit{Coherence} attribute when it is the N-th attribute predicted}.}\label{tab:nth_element_performance}
\end{table}

\subsection{Self-Inconsistency}
\label{subsec:self-consistency}

\begin{figure*}[h]
   \centering
   \includegraphics[width=\textwidth]{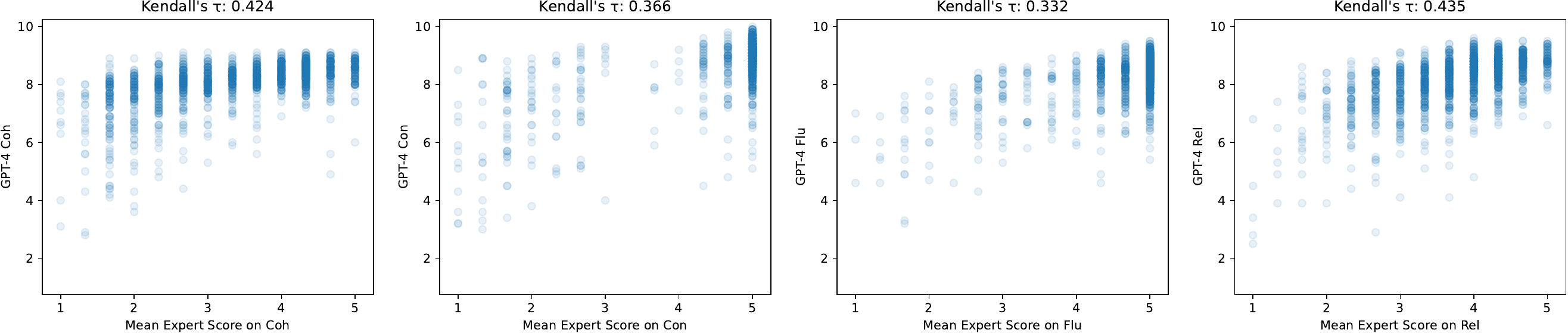}
   \caption{
       \textbf{Scatter-plots of evaluated score versus expert judgements} reveal that while many papers claim 0.40 $\tau$ is strong performance, the correlation with human judgements still needs substantial improvements.
       Even with correlation of over 0.40 Kendall's $\tau$, we notice that any individual evaluation may lie within a very wide range as compared to the ground-truth labeled by experts.
       Note that the full range of 1-10 is underutilized again.
       }
    \label{fig:scatter}
\end{figure*}

The general performance and self consistency of LLM-based metrics is problematic when considering actual uses.
While \citet{stureborg-etal-2024-characterizing} point out that even low correlations with human judgements can be used to make high-confidence comparisons on the system level, correlation needs to be very high for any individual prediction by the automatic evaluator to be trusted.
\Cref{fig:scatter} shows scatter plots of predictions made by an LLM evaluator as compared to human judgements.
It shows that even predictions on a single example can vary widely by the same model given slight prompt modifications or even just sampling at temperature settings $>0$.

To analyze the self-inconsistency of LLM evaluators, we calculated inter-sample agreement using Krippendorff’s $\alpha$. \Cref{tab:krippendorff_alpha_summary} shows that the self-consistency is worse than the consistency between multiple human annotators.

\begin{table}[h]
        \centering
        \small
        \begin{tabular}{r|r}
        \toprule
              &  {\bf $\alpha$} \\\midrule
        Inter-annotator agreement (Human) &  0.659 \\ \midrule
        Inter-sample agreement (GPT-4)    &  0.587 \\
        \bottomrule
    \end{tabular}
    \caption{{\bf Krippendorff’s $\alpha$ for inter-annotator agreement (Human) and inter-sample agreement (GPT-4).}
    \label{tab:krippendorff_alpha_summary}}
\end{table}

\subsection{Sensitivity to Temperature and CoT}
\citet{chiang-lee-2023-closer} determined that CoT is not always helpful in improving the performance of LLM-evaluation.
We investigate this further by tuning temperature settings on the task under CoT and non-CoT approaches.

Many guidelines for LLM prompt-engineering have unintuitive implications when combined.
Generally, lower temperature generations are preferred during simple inference tasks with LLMs.
Also, Chain-of-Though (CoT) is a popular \citep{CoT_prompting} strategy to increase text generation quality, reasoning, and task performance across many settings.
However, we find that when using CoT prompting, lower temperatures are not preferable.
This result is not immediately obvious.
Instead, we propose using multiple generations at higher temperatures.
Looking through the raw outputs, this seems to be due to a more diverse set of explanations that lead to a more robust numerical prediction.
We posit this is similar to combining many weak estimators, and that increasing temperature helps decrease the correlation between each estimators prediction.

\begin{figure}[h]
   \centering
   \includegraphics[width=.48\textwidth]{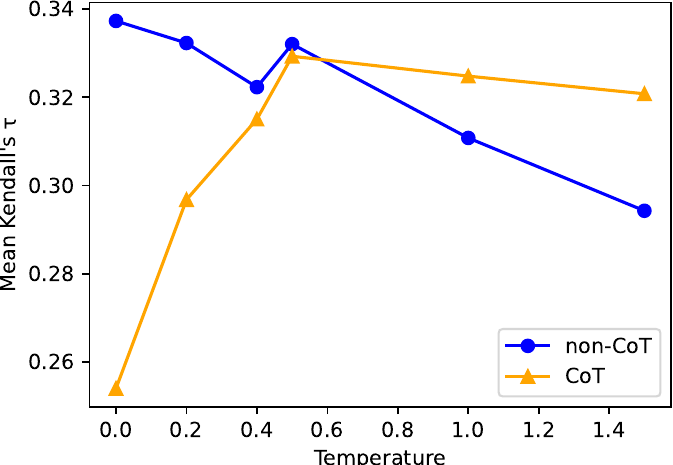}
   \caption{
       \textbf{Performance of CoT and non-CoT prompting at varying Temperatures}.
       Each prediction is computed by the average of 10 generations.
       Low temperatures are beneficial when making simple predictions, but higher temperatures (to a point) help improve performance when using Chain-of-Thought (CoT) prompting.
       This could be because of a more diverse set of explanations, leading to more unique features for prediction.
    }
    \label{fig:cot_temperature}
\end{figure}

\Cref{fig:cot_temperature} shows that CoT prompting benefits from higher temperatures, while non-CoT performs better with lower temperatures.
Setting outputs to deterministic generation (temperature of 0) may serve counter-productive since generating the most likely token ensures granularity is limited by the original range of the scoring. 

We produce predictions at various temperatures using GPT-3.5-Turbo in \Cref{fig:cot_temperature}, showing that increasing temperature steadily reduces performance of non-CoT prompts, while performance of CoT prompts increases sharply until approximately 0.5.
CoT prompts performance then subsequently drops off or plateaus as temperature is increased further.
This trend is not just over the average scores on SummEval.
\Cref{fig:cot_temperature_by_attr} shows similar plots for both non-CoT and CoT prompts, plotting the performance on each label in SummEval individually.
The trends described in \Cref{fig:cot_temperature} seem to replicate on each label in this dataset.
Our findings show that a single generation at temperature 0 outperforms the best tuning of multi-sample CoT is cheaper and simpler than the weighted average approach from \citet{GEval}.
When using CoT, our results motivate drawing multiple samples while tuning temperature appropriately to maximize performance.

\subsection{Sensitivity to Source Document}
While the long-context abilities of LLMs allow predictions over more complex documents, we find that the model's use of the provided source document (the article being summarized) is questionable during automatic evaluation.
The presence of this source document substantially affects ratings on fluency, which should be independent of the article text.
The table below shows performance drops of LLM-based evaluation using GPT-3.5-Turbo when removing the Source document, although many of the categories which surely require the document to render a sensible judgement remain relatively high-performing.
The LLM-evaluator may be picking up on spuriously correlated features when predicting its judgement, indicating a potentially problematic bias.

\begin{table}[h]
        \centering
        \small
        \begin{tabular}{rrrrrr}
        \toprule
        \textbf{Source Doc} & Coh & Con & Flu & Rel & \textbf{Avg} \\
        \midrule
        Included    &  \textbf{.346}  & \textbf{.250} & \textbf{.237} & \textbf{.330} & \textbf{.291} \\
        Excluded    &  .291  & .167 & .212 & .183 & .213 \\
        $\Delta$    & -.055  & -.083 & -.025 & -.147 & -.078 \\
        $\%\Delta$  & -15.9  & -33.2 & -10.6 & -44.6 & -26.7 \\
        \bottomrule
    \end{tabular}
    \caption{\textbf{Performance of GPT-3.5-Turbo with and without Source Document}.
    Removing the source document (unsurprisingly) substantially reduces the performance of the automatic evaluator.
    However, this is also true for attributes that should not be dependent on the source document in the first place, such as Fluency. 
    For categories such as relevance, making a prediction on the summary quality without the article should be impossible.
    }    
    \label{tab:granularity_methods}
\end{table}

Overall performance drops by 27\% (relative), heavily driven by a drop in performance on relevance.
While relevance is a dimension of evaluation that depends entirely on the source documents match with the summary, GPT-3.5-Turbo is able to find features that may be correlated with the expert scoring.

\section{Case Study}
\label{sec:case_study}

Using the lessons learned from SummEval in \Cref{sec:analysis}, we determine a few simple guidelines to significantly improve automatic evaluation with LLMs (see \Cref{tab:solutions}).
We evaluate whether these guidelines improve performance by comparing to two previous works: G-Eval \citep{GEval} and a follow-up work by \citet{chiang-lee-2023-closer}.
\citet{chiang-lee-2023-closer} establish SOTA performance on SummEval, beating G-Eval's correlation with human judgements on the dataset.
However, some \citep{bhandari-etal-2020-evaluating, liu-etal-2023-revisiting} have pointed out issues in these style of datasets, including that (1) expert ratings themselves include a lot of disagreement, (2) closed-source LLMs may have been trained on these well-established datasets, and (3) conclusions on these datasets don't always hold for new systems.

For these reasons, we evaluate our system on RoSE, a summary evaluation dataset built carefully in a multi-stage process to maximize label quality and is unlikely to be included in GPT training data.
RoSE's target label is the metric \textit{Atomic Content Units} (ACU) which is a normalized metric ranging from 0 to 1.
Note that the CNNDM partition of the dataset is shared with SummEval, meaning that performance on this data is an in-domain test, while the other two partitions of RoSE serve as out-of-domain tests.

\paragraph{Implementation of Previous Work}
\citet{chiang-lee-2023-closer} point out issues in replicating the reported correlation values from the G-Eval paper.
Therefore, we compare with these works by re-implementing their systems using the descriptions in their methods and released code, and compute all correlation values from scratch.
Both \citet{chiang-lee-2023-closer} and G-Eval were approaches designed for OpenAI's Completions API endpoint, as opposed to a ChatCompletion end-point, which is more limited in formatting and has no access to token probabilities.
We map the prompts into a Chat format by simply placing them into the user prompt.\footnote{Experiments with using our own prompt this way indicated a small but not statistically significant performance increase.}
For G-Eval, we sample 10 times and average the score to approximate their expected value calculation (which was done by multiplying token probabilities extracted from the model). 
We use auto-CoT as specified, but notice that this causes a higher proportion of ``failed'' generations which give texts but omit any final, parseable score.
\citet{chiang-lee-2023-closer} suggest not including auto-CoT or any evaluation steps in their approach.
For our method, we include the evaluation steps undergone by annotators for the ACU metric.
This text is taken directly from \citet{liu-etal-2023-revisiting} with edits only for grammar and conciseness.
Finally, we use the rate-explain setting they describe since it is one of their two best settings. They state rate-explain and analyze-rate are ``do not see rate-explain to be significantly better (or worse) than analyze-rate''. While the authors don't point this out, rate-explain is much cheaper and faster for generation given you can safely stop generation after the rating has been produced.

We compare all methods on GPT-4-Turbo. Our method, as determined by insights from \Cref{sec:analysis}, relies on a 1-10 scoring granularity and includes both evaluation steps and a definition of ACU (which is copy pasted from \citet{liu-etal-2023-revisiting} and also added to other two approaches). We use non-CoT prompting at a temperature of 0, and generate a single output. \Cref{tab:solutions} summarizes these approaches.
None of these parameters are tuned on RoSE.
While each solution in \Cref{tab:solutions} might look commonly used techniques, to the best of our knowledge, none of existing work has combined them into a single recipe and conduct an empirical study to verify the effectiveness of the techniques.

    \begin{table*}[h]
        \centering
        \small
        \begin{tabular}{ll}
        \toprule
        \textbf{Issue w/ LLM evaluators} & \textbf{Reasonable Approach to Mitigate} \\
        \midrule
        Low granularity for distinguishing summaries & Widen scores to 1-10 star scale \\
        CoT prompting requires tuning temperature & Remove CoT and set temperature to 0  \\
        Removing source document impacts performance & Keep source even for attributes which don't require it \\
        Multi-attribute labels are highly correlated & Predict only one attribute per generation \\
        \bottomrule
    \end{tabular}
    \caption{\textbf{Identified issues have immediate and actionable mitigations}
    }
    \label{tab:solutions}
\end{table*}

\paragraph{Results}
Our method outperforms both \textit{G-Eval} and \textit{rate-explain} on the CNNDM and SAMSum partitions.
The performance of our method achieves Kendall's $\tau = 0.220$ on the in-domain test set, and $\tau = 0.308$ on SAMSum, indicating this partition may be easier to evaluate.
While we outperform \citet{chiang-lee-2023-closer} on SAMSum, the difference is not statistically significant.
This significant variation in performance is due to prompting strategies, indicating a lot of room for performance improvements by closer studies in prompt engineering.

\begin{figure}[h]
   \centering
   \includegraphics[width=.48\textwidth]{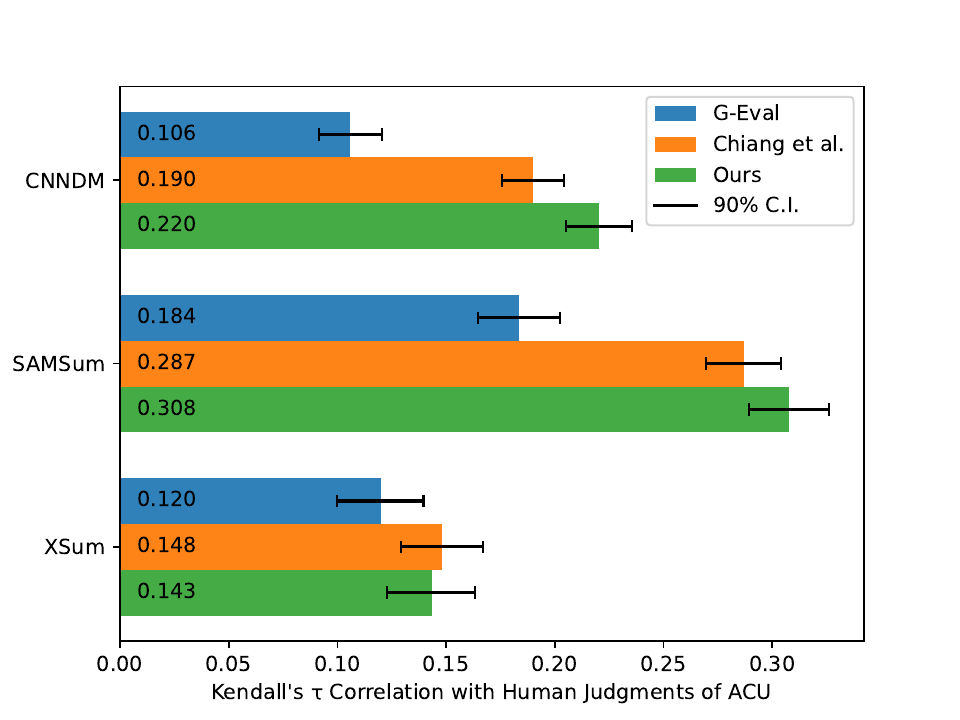}
   \caption{
       \textbf{Performance Comparison on the RoSE benchmark}.
       Our approach performs statistically significantly better than the SOTA LLM-evaluator for summarization \citep{chiang-lee-2023-closer} on the CNNDM dataset partition, and significantly better than G-Eval on both CNNDM and SAMSum. 
       Confidence intervals are computed through bootstrap sampling.
    }
    \label{fig:ours_vs_geval}
\end{figure}

\section{Related Work}
\label{sec:related_work}
Automatic evaluation has been dependent on human annotations. Traditional automatic evaluation metrics such as ROUGE~\citep{lin-2004-rouge}, BLEU~\citep{papineni-etal-2002-bleu}, and METEOR~\citep{banerjee-lavie-2005-meteor} consider token-level n-gram matching between system outputs and reference texts. Later, embedding-based automatic evaluation such as BERTScore~\citep{zhang2019bertscore}, BLEURT~\citep{sellam-etal-2020-bleurt}, and MoverScore~\citep{zhao-etal-2019-moverscore} were developed to take the semantic similarity into account.
Extensive efforts to remove the reliance on manually written reference texts have been attempted by creating reference-free automatic evaluation metrics~\citep{louis-nenkova-2013-automatically,fonseca-etal-2019-findings,scialom-etal-2019-answers,scialom-etal-2021-questeval,vasilyev-etal-2020-fill,rei-etal-2021-references}. However, \citet{deutsch-etal-2022-limitations} have pointed out the current limitations as the measures of how well models perform a task.

Following the line of work, recent studies in LLM evaluators have shown that LLMs can be high-quality evaluators for various NLP tasks~\citep{fu2023gptscore} including Summarization~\citep{chen2023exploring,wang2023chatgpt,liu2023geval,gao2023humanlike,shen2023large,wu2023long}, Machine Translation~\citep{kocmi2023large}, Factual Consistency Evaluation~\citep{luo2023chatgpt}, and other text generation tasks~\citep{chen2023exploring,wang2023chatgpt,chan2023chateval,kasner2024referencebased}.

However, they have primarily focused on improvements through prompt engineering.
Among them, only a few studies have tried to reveal the limitations of LLM evaluators. 
They have reported that LLM evaluators have position bias---a preference for the first example of a pairwise comparison~\citep{wang2023large,zheng2023judging}; verbosity bias--preference for longer texts~\citep{zheng2023judging,wu2023style}; and self-enhancement bias---LLM evaluators prefer text generated by themselves~\cite{zheng2023judging,panickssery2024llmevaluators}.
\citet{koo2023benchmarking} have reported cognitive biases in LLM evaluators.
Following the studies, our paper aims to dig deeper to share quantitative analysis on these points and beyond. 
Our work partially overlaps with the recent work by \citet{ohi2024likelihoodbased}, who studies likelihood bias in LLM evaluators across data-to-text and grammatical error correction tasks. However, our work differs in that we use a different metric (i.e., perplexity) to assess the bias and focus on a different target task (i.e., summarization), providing a new perspective on this issue.

\section{Conclusion}

We have provided a series of analyses into biased and inconsistent behaviors exhibited by LLM evaluators for the task of text summarization.
Our findings show that (1) LLM evaluators are disproportionately biased towards low perplexity summaries than is helpful (familiarity bias), (2) they fail to respect scoring scales given to them when attempting to increase the granularity of scores (score bias), (3) they show degradation in multi-attribute judgment, being influenced by their previous ratings (anchoring effect).
They are inconsistent their own judgements depending on settings such as inclusion of source documents.

In attempts to solve some of these issues, we share a recipe to mitigate these issues and show that we are able to significantly outperform the current SOTA method for LLM-based summary evaluation on the CNNDM partition of RoSE 90\% confidence.
Our work suggests that more effort should be allocated towards understanding and remedying the issues exhibited by LLM evaluators.

\section*{Limitations}

\textit{Reliance on GPT-based models}.
We experiment primarily on GPT-based, proprietary models from OpenAI due to their SOTA performance on automatic evaluation of text summarization.
However, this means it is unclear how well our results generalize to other LLMs such as Llama-2, Vicuna, Alpaca, etc.
Do to constraints in time and budget, extending the analysis to investigate other LLMs was not possible during the time this work was carried out.
This project involved generating more than 560,000 outputs from OpenAI models; repeating the experiments on several models amounts to substantial effort and resources.
Future work could aim to replicate and extend our analysis to further models.

\textit{Reliance on SummEval for analysis}.
Our analysis section primarily investigates issues by measuring performance of various model and prompt configurations against SummEval.
There is a risk that our results to do generalize well beyond 
For this reason, we also sought to measure performance on the RoSE benchmark, which is comprised of three datasets in different domains.
We find that addressing the issuess seen in SummEval significantly improves performance on one of the domains, and has insignificant but positive results on the other domains.

\textit{Limited solutions}.
Although we investigate solutions to some of the identified issues in this paper, many remain to be studied and may provide the research community with directions for future research efforts.
LLM's inconsistencies and biases as automatic evaluators is tough to build solutions around.
There is ample opportunity for creative solutions, and while our work offers some, its main focus is in identifying the existing issues in the first place.

\section*{Ethics Statement}
As this study focuses on text summarization and uses publicly available datasets, we do not see any clear ethical implications or considerations.
We adhere to ethical research practices.

\bibliography{anthology,custom}
\bibliographystyle{acl_natbib}

\appendix
\onecolumn

\section{Our Method for Prompting LLM Evaluation}
\label{sec:our_best_method}

\begin{figure}[h]
    \centering
    \begin{lstlisting}
Document:
{{article}}

Summary:
{{summary}}

Evaluation Form (Scores ONLY):
{{metric}}:
    \end{lstlisting}
    \caption{\textbf{User text input, used in conjunction with the system prompt in \Cref{fig:system_prompt}}.
    The next immediate token is expected to be within the range of 1 to 10, but oftentimes the models will output restatements of the metric name or other content first.
    In general, we find it is safe to stop the model generation after 10-20 tokens and parse this output using regex to find the first digit.
    }
    \label{fig:user_prompt}
\end{figure}

\textbf{System Prompt:}
\begin{lstlisting}
    You are the automatic summary evaluator of a writing editor:
    - You consider an input document and a corresponding summary
    - You evaluate the summary according to one important quality:
        1. ACU Salience (1-10) - a desired summary quality that requires the summary to include all and only important information of the input article. Salience can be determined by dissecting the summaries into fine-grained content units and defining the annotation task based on those units. Specifically, we introduce the Atomic Content Unit (ACU), elementary information units which no longer need to be further split for the purpose of reducing ambiguity in human evaluation. The evaluation process is decomposed into extracting facts from one text sequence, and checking for the presence of the extracted facts in another sequence.
    - All ratings are between 1-10 where 1 is very poor and 10 is very good.
    - Your evaluation should be critical and careful, and should closely match the ratings of experts. This evaluation is very important.
    - Consider these aspects when evaluating:
        1. ACU Writing - Read the document carefully and identify all Atomic Content Units (ACUs) and facts.
        2. ACU Matching - Read the summary and compare it to the list of ACUs. Check what proportion of the extracted ACUs that the summary correctly covers.
        3. Assign a score for ACU Salience on a scale of 1 to 10, where 1 is the lowest (covers very few of ACUs) and 10 is the highest (covers all important ACUs) based on the Evaluation Criteria.
    
    The user will give you both the article (document) and summary, and prompt you to provide an evaluation. Respond with your integer 1-10 score first, then a rationale.
    
    Example:
\end{lstlisting}

\textbf{User Prompt:}
\begin{lstlisting}
    Document:
    {{article}}
    
    Summary:
    {{summary}}
    
    Evaluation Form (Scores ONLY):
    ACU Salience:
\end{lstlisting}

\section{Familiarity Bias}

\begin{figure}[H]
   \centering
   \includegraphics[width=.48\textwidth]{figures/ppl_gpt4_by_davinci.pdf}
   \includegraphics[width=.48\textwidth]{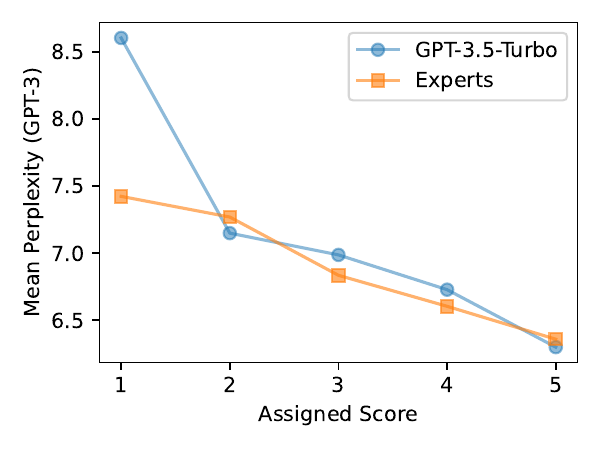}
   \caption{
       \textbf{Average Perplexity associated with Automatic Evaluation Score, for each Attribute}. While GPT-4 shows a preference for low perplexity samples, GPT-3.5-Turbo seems to show a dis-preference for high perplexity examples.
     }
    \label{fig:alpaca_ppl}
\end{figure}

\section{Anchoring Effect in Multiple Judgments}
\label{apx:prob_x_given_y}
\begin{figure}[H]
   \centering
   \begin{subfigure}{0.33\linewidth}
     \centering
     \includegraphics[width=\textwidth]{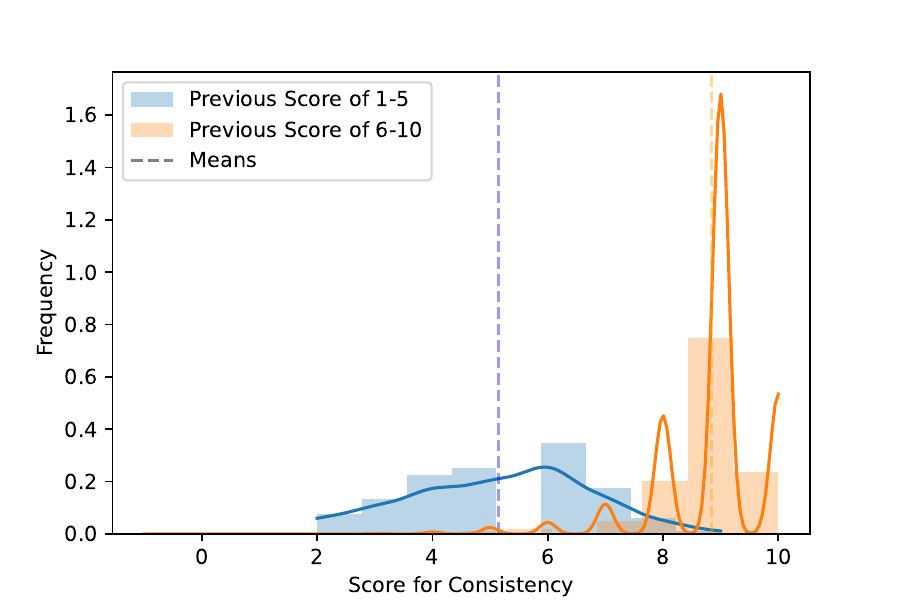}
     \caption{Consistency conditioned on coherence}
   \end{subfigure}
   \begin{subfigure}{0.33\linewidth}
     \centering
     \includegraphics[width=\textwidth]{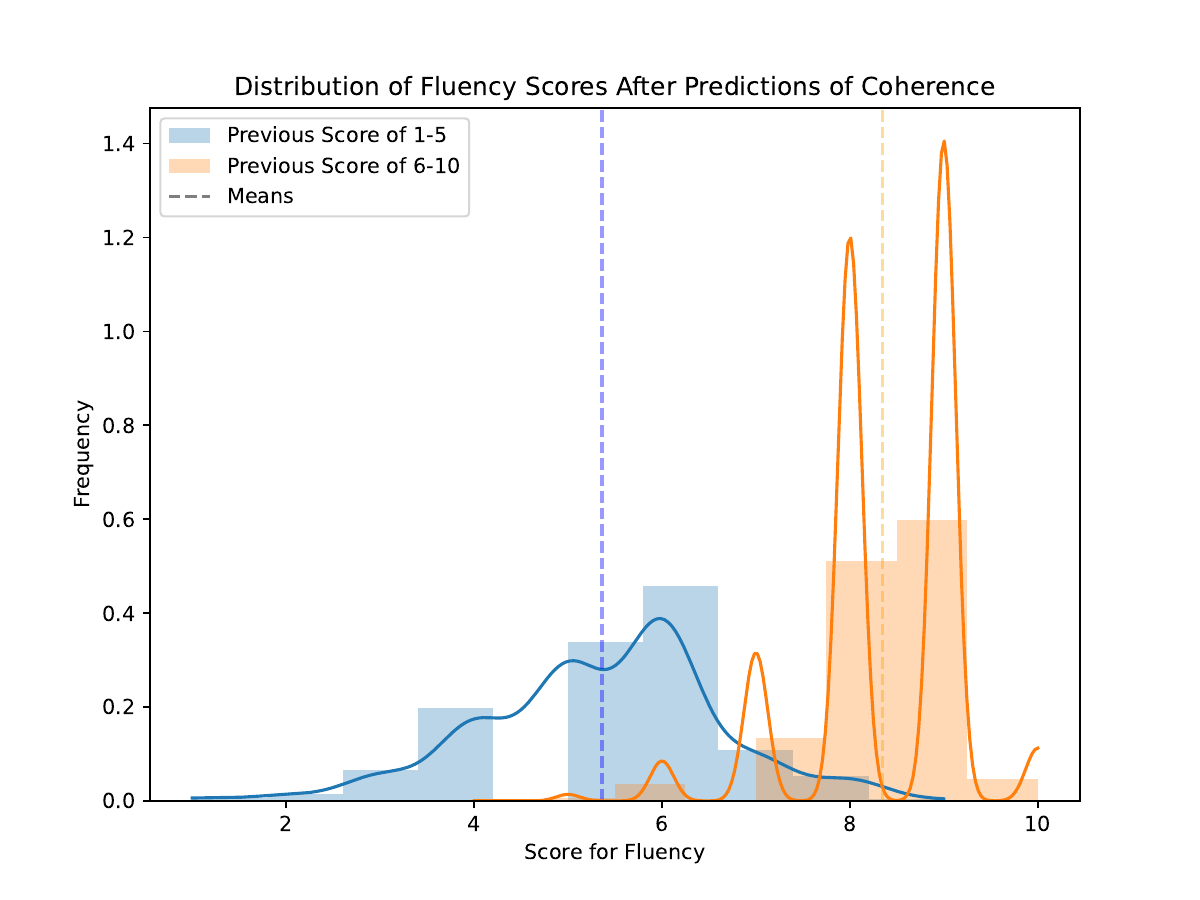}
     \caption{Fluency conditioned on coherence}
   \end{subfigure}
   \begin{subfigure}{0.33\linewidth}
     \centering
     \includegraphics[width=\textwidth]{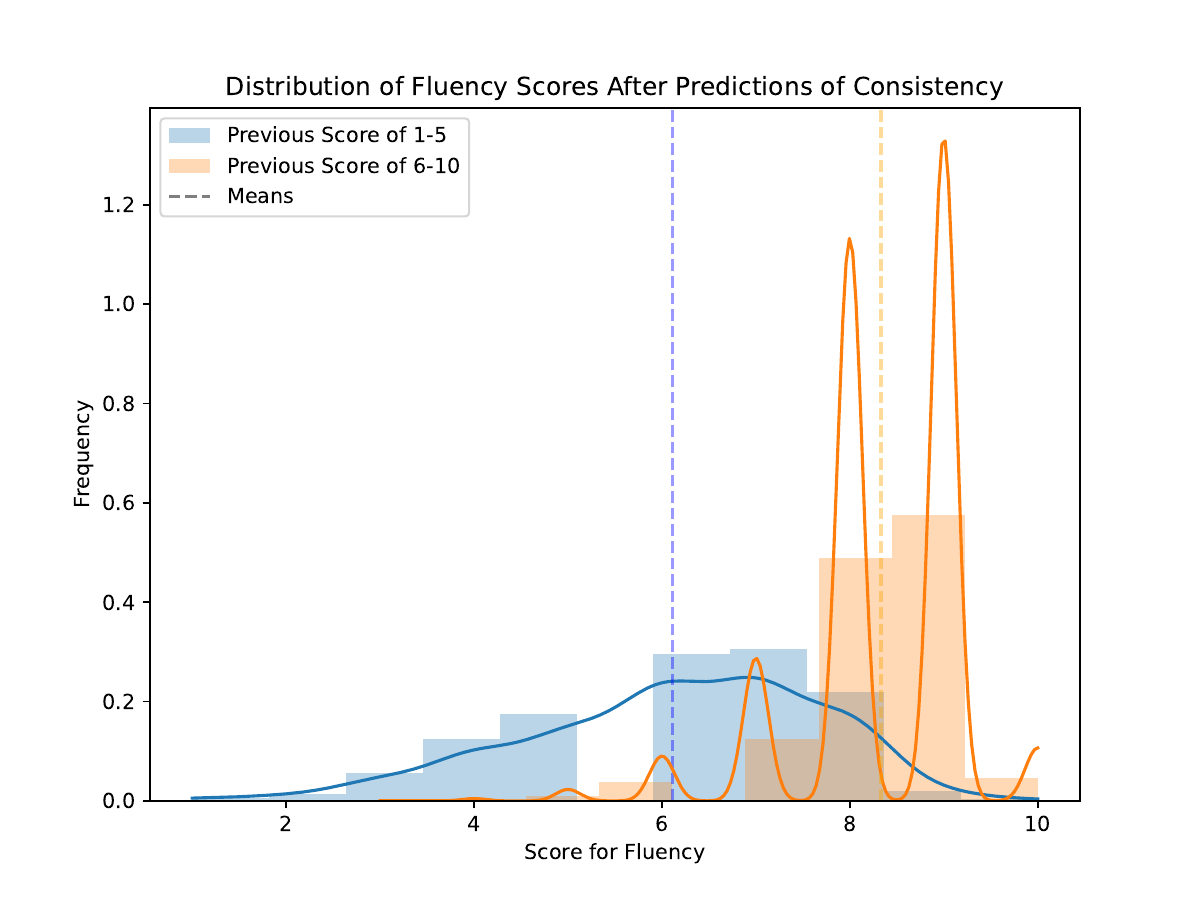}
     \caption{Fluency conditioned on consistency}
   \end{subfigure}
   \begin{subfigure}{0.33\linewidth}
     \centering
     \includegraphics[width=\textwidth]{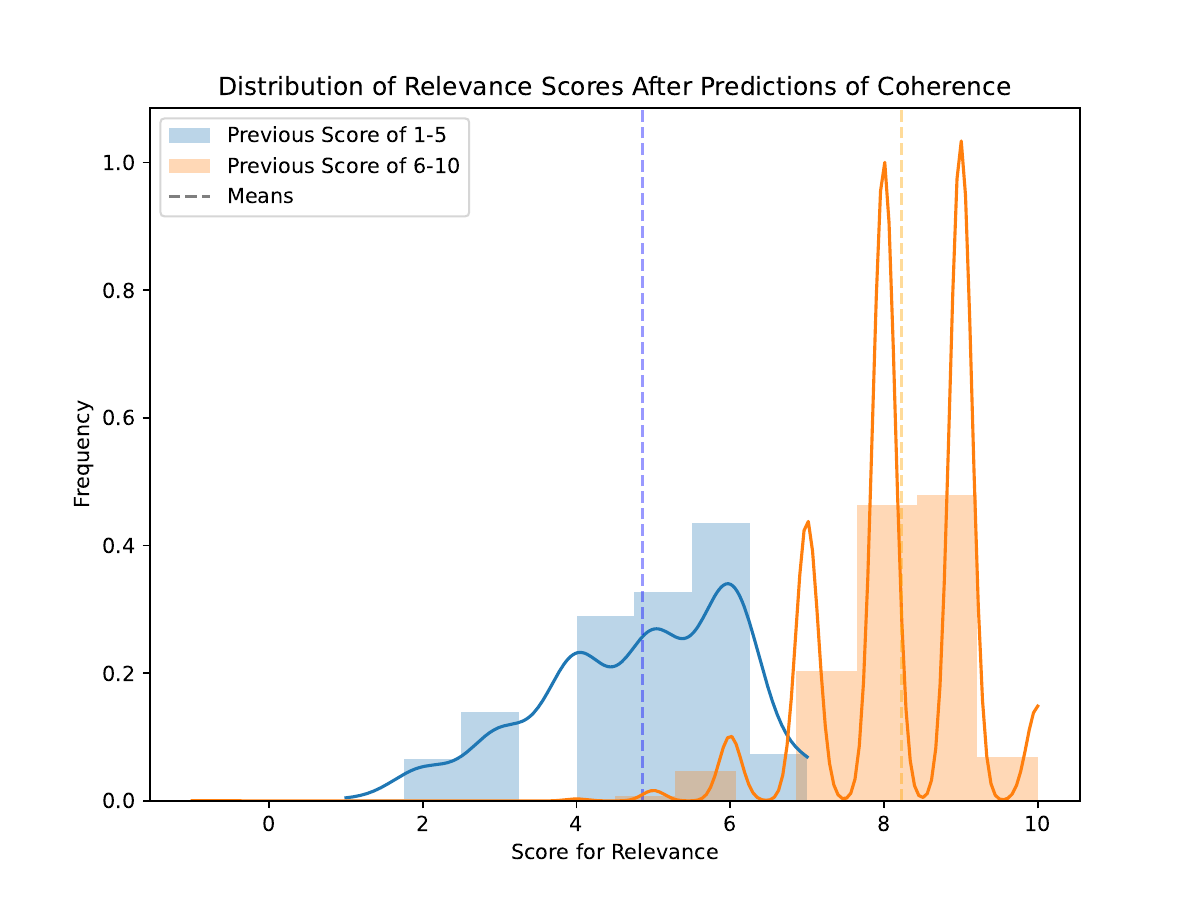}
     \caption{Relevance conditioned on coherence}
   \end{subfigure}
   \begin{subfigure}{0.33\linewidth}
     \centering
     \includegraphics[width=\textwidth]{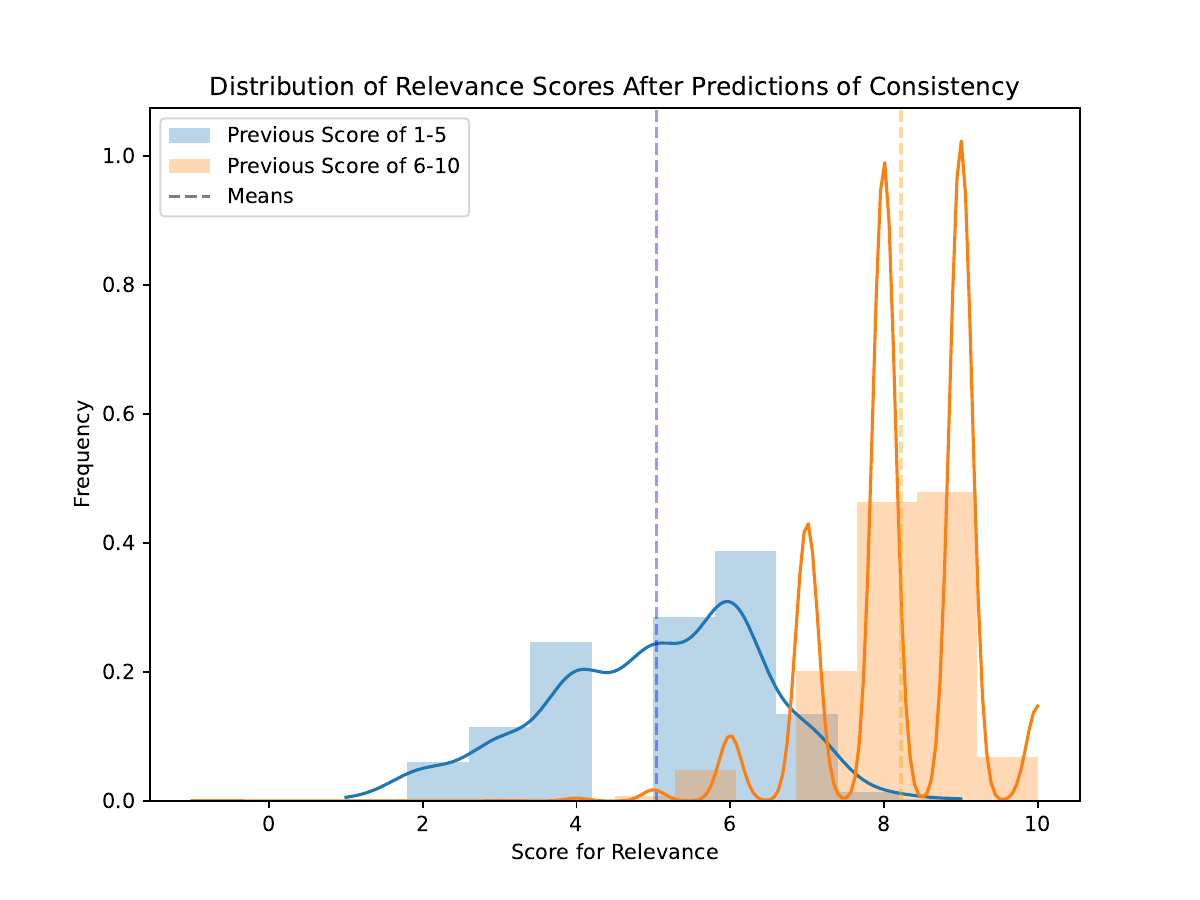}
     \caption{Relevance conditioned on consistency}
   \end{subfigure}
   \begin{subfigure}{0.33\linewidth}
     \centering
     \includegraphics[width=\textwidth]{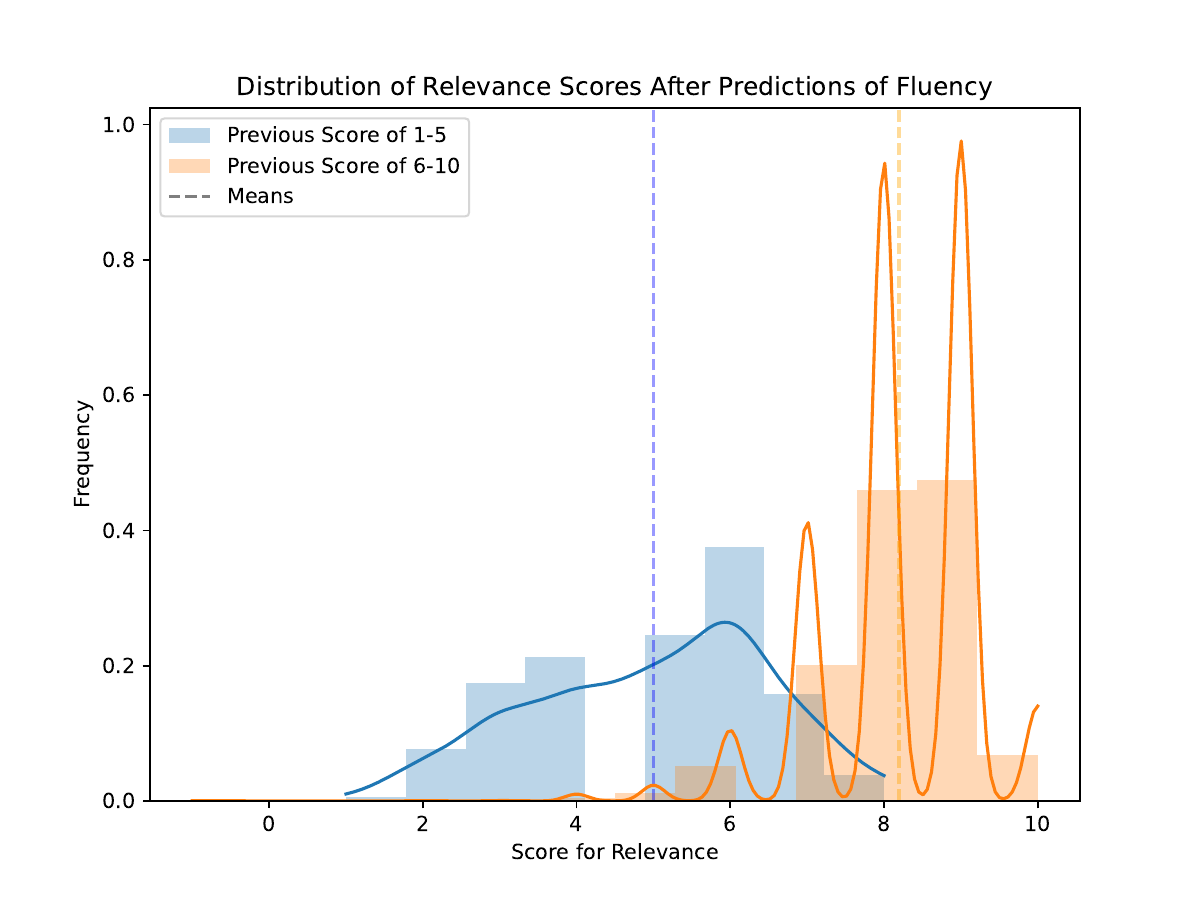}
     \caption{Relevance conditioned on fluency}
   \end{subfigure}
   \caption{
       \textbf{Distribution of scores conditioned on the values of previous scores in the same generation}.
    }
\end{figure}

\section{Performance of CoT and non-CoT prompting for each attribute}
\begin{figure*}[h]
   \centering
   \includegraphics[width=\textwidth]{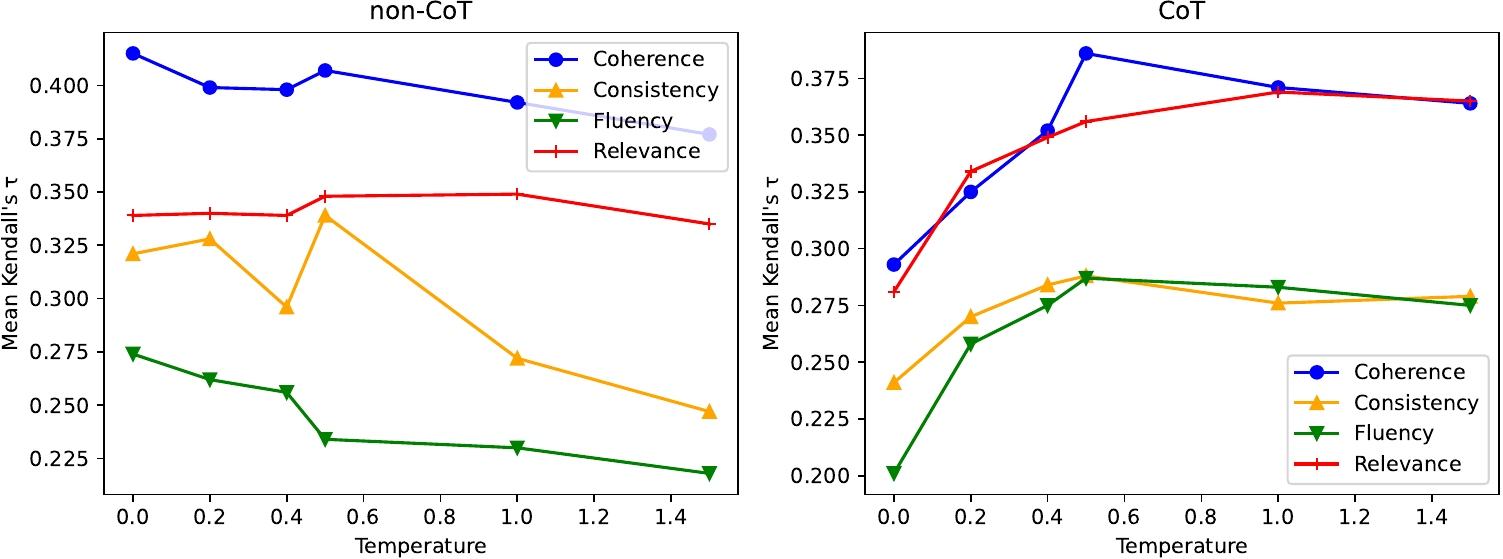}
   \caption{
       \textbf{Performance of CoT and non-CoT prompting at varying Temperatures}.
       Trends shown in \Cref{fig:cot_temperature} hold up on individual target dimensions in SummEval.
       In general, non-CoT predictions are harmed by higher temperatures, while CoT predictions are improved (though with diminishing returns).
    }
    \label{fig:cot_temperature_by_attr}
\end{figure*}

\section{Self-inconsistency}

\begin{table}[th]
        \centering
        \small
        \begin{tabular}{r|r|rrrr|r}
        \toprule
              &   & {\bf Coh} & {\bf Con} & {\bf Flu} & {\bf Rel} & {\bf Avg} \\\midrule
        Human & Inter-annotator agreement & 0.559 & 0.899 & 0.726 & 0.453 & 0.659 \\ \midrule
        GPT-4 & Inter-sample agreement    & 0.646 & 0.630 & 0.484 & 0.589 & 0.587 \\
              & Single- vs Multi-attribute & 0.493 & 0.667 & 0.472 & 0.421 & 0.513 \\
              & 1-5 star vs 1-10 score    & 0.731 & 0.442 & 0.506 & 0.707 & 0.597 \\
              & 1-5 star vs 1-100 score   & 0.557 & 0.128 & 0.471 & 0.600 & 0.439 \\
        \bottomrule
    \end{tabular}
    \caption{{\bf Full results for Krippendorff’s $\alpha$ values for inter-annotator agreement and self-consistency evaluation.} For GPT-4's ``inter-sample'' agreement, multiple samples are considered annotations made by different annotators. For the remaining Krippendorff’s $\alpha$ values were calculated for the agreement between judgments obtained for different settings (e.g., using single-attribute template and multi-attribute template.) For 1-10 and 1-100 score judgments, the judgements were converted into the same scale (1-5) by binning the numbers.}
    \label{tab:krippendorff_alpha}
\end{table}

\end{document}